# Evolutionary Multi-Objective Optimization Algorithm for Community Detection in Complex Social Networks

**Shaik Tanveer Ul Huq[,2], Vadlamani Ravi[1*], Kalyanmoy Deb[3]**

[1]Institute for Development and Research in Banking Technology,
Castle Hills Road No. 1, Masab Tank, Hyderabad-500057, India

[2]SCIS, University of Hyderabad, Hyderabad – 500040, India

[3]Koeing Endowed Chair Professor, Department of Electrical & Computer Engineering,
Michigan State University, East Lansing, MI, USA

shaiktanveerulhuq@gmail.com; vravi@idrbt.ac.in; kdeb@egr.msu.edu

**Abstract**

Most optimization-based community detection approaches formulate the problem in a single or bi-objective framework. In this paper, we propose two variants of a three-objective formulation using a customized non-dominated sorting genetic algorithm III (NSGA-III) to find community structures in a network. In the first variant, named NSGA-III-KRM, we considered *Kernel k means*, *Ratio cut*, and *Modularity,* as the three objectives, whereas the second variant, named NSGA-III-CCM, considers *Community score*, *Community fitness* and *Modularity,* as three objective functions. Experiments are conducted on four benchmark network datasets. Comparison with state-of-the-art approaches along with decomposition-based multi-objective evolutionary algorithm variants (MOEA/D-KRM and MOEA/D-CCM) indicates that the proposed variants yield comparable or better results. This is particularly significant because the addition of the third objective does not worsen the results of the other two objectives. We also propose a simple method to rank the Pareto solutions so obtained by proposing a new measure, namely the ratio of the hyper-volume and inverted generational distance (IGD). The higher the ratio, the better is the Pareto set. This strategy is particularly useful in the absence of empirical attainment function in the multi-objective framework, where the number of objectives is more than two.

## 1. Introduction

A Complex network can be considered as a graph, having set of a nodes and edges between them. Examples of such networks are The World wide web, collaboration networks, online social networks, Food Web, biological networks etc.

Analysis of these complex networks provides us better insights into the quality of interconnections among the nodes such as the identification of important nodes and the structure of underlining communities present in it. Community detection is paramount having numerous applications in e-commerce, communication networks social networks, biological systems, health care, economics, academia, fraud detection etc. [1].

The issue of detecting communities is to find the sets of nodes such that, each set has nodes that are thickly connected with one another and are loosely connected with the nodes present in the remaining sets. This problem is NP hard [1]. In the last decade, numerous approaches have been propounded to find communities in networks. Some of the techniques are hierarchical clustering algorithms, graph partitioning methods and evolutionary algorithms.

In this paper, community detection in a given undirected and unweighted network is formulated as a multi-objective optimization problem with three objectives and is solved using NSGA-III [2]. Throughout this paper, the words community and cluster are used interchangeably.

In what follows, section 2 presents the related work, section 3 presents the motivation, section 4 describes the contribution of the present study, section 5 presents basic definitions, section 6 presents proposed methodology, section 7 describes the datasets analyzed, section 8 displays results obtained and discussion thereof and finally, section 9 concludes the paper.

2. Literature Survey

In the last decade, several meta heuristic algorithms have been suggested to solve community detection problem in complex networks. In 2003, Newman introduced a classical [3] algorithm which optimizes Modularity in a greedy manner. It uses agglomerative hierarchal clustering method to iteratively maximize Modularity. Later in 2008, Blondel et al. designed another classical [4] two-phase algorithm, which also optimizes Modularity. In the first phase, nodes in one community are shifted to another community one at a time iteratively, if Modularity increases and in second phase communities are merged to get larger communities. In the same year, Pizzuti proposed GA-NET [5]. It uses locus-based representation to represent a community structure and optimizes Community score to identify communities in a network. Thereafter, in 2011, Gong et al. developed MEME-NET [6]. It is observed that Modularity suffers from resolution limit problem [7]. So, they optimized Modularity Density instead of Modularity using genetic algorithm (GA) and including hill climbing for local search to find communities in a network Later In 2012, Shang et al. proposed MIGA [8]. It also optimizes Modularity using GA and included simulated annealing to perform local search to find communities in a given network. Then, Pizzutti introduced MOGA-NET [9]. It optimizes two objective functions viz., Community score and Community fitness using GA to detect communities in a network. Then, In 2014, Gong et al. developed MODPSO [10]. It optimizes two objective functions viz., kernel k means and Ratio cut using discrete particle swarm optimization algorithm to find communities in a network. This approach can be used for both signed and unsigned networks. Later, in 2017, Abdollahpouri et al. proposed MOPSO-Net [11], a customized version of particle swarm optimization by altering the moving technique of particles. While moving from one iteration to another, this method uses Normalized mutual information (NMI). NMI needs the ground truth cluster structure of the graph as input. Hence, this method is not helpful if we do not know the ground truth community structure of the network in advance. In 2018, Yuanyuan et al. proposed two quantum inspired evolutionary algorithms viz., QIEA-net and iQIEA-net [12] to find community structures. QIEA-net detects the communities by optimizing Modularity, and in IQIEA-net, it takes the help of the classical partitioning algorithm. Most recently, Tahmasebi et al. [13] in 2019 proposed a many-objective community detection algorithm which takes five objectives. Out of the five, two objectives cannot be calculated if the ground truth community structure is unknown which is indeed the case in real-life problems. In such cases, those methods cannot be used because the very task there is to find communities in the conspicuous absence of ground truth.

To sum up, single objective community detection algorithms lead to some difficulties such as limiting to particular community structure properties. Then, bi-objective formulations did indeed leave out some important measures, which could potentially be used as objective functions. We noticed that some of the measures are indeed non-overlapping conceptually. They describe different aspects of a community. Hence, a different approach is proposed in the current paper, which is a multi-objective (three objective) optimization framework in two variants to search for communities in complex social networks. This is a clear departure from all the works appeared in the literature so far.

## 3. Motivation

To the best of our knowledge, except for one of the latest pappers, all the works in the literature, formulated community detection of networks as an optimization problem in either single objective or two objectives. Frameworks that considered single objective have considered mostly Modularity as the objective function, while those with two objectives considered two objectives as follows: Kernel k means & Ratio cut or Community fitness & Community score or Ratio cut & ratio association or Modularity (by dividing the Modularity into two parts and considered each part as one objective). In bi-objective optimization frameworks, one objective maximizes the density of communities and the other minimizes the fraction of interlinks present between communities in the network. (For instance, Kernel k means tries to find the solution with maximum community density and Ratio cut tries to find the solutions with minimum fraction of interlinks between communities). For evaluating the effectiveness, they employed Modularity and NMI (for networks with known the ground truth communities) as external measures outside the optimization process.

If we consider only two objectives, we may get solutions having high community density and less interlinks between communities. However, these solutions may or may not have good community structure. For example, in a network N, if we consider a solution with only one community consisting of all the nodes in the network, that solution has maximum intra-links and zero interlinks but it may not the best structure because the Modularity value becomes zero for that solution and it does not satisfy the goal of the problem namely to find distinct, non-overlapping communities.

Most recently, Tahmasebi et al. [13] also proposed a many-objective community detection algorithm which takes five objectives. Out of five, two objectives cannott be calculated if the ground truth community structure of the given network is unknown. Thus, in effect, it reduces to three-objective formulation.

Further, they used another objective function Coverage and mentioned that Coverage is the proportion of edges inside the community to the total edges in network. Thus, it refers to the density of a given cluster.

In this paper, we propose a multi-objective optimization framework using three objectives, which try to find solutions with good community densities, less fraction of interlinks and good community structures as well. Our approach is more generic enough as it does not need to know the ground truth community structure in advance. Toward this end, we employed customized NSGA-III as the optimizer.

## 4. Contributions

- Some studies [11] performed the selection of solutions after every generation based on NMI. But, it should be noted that computation of NMI requires the ground truth community structure. These methods are not helpful if we do not know the ground truth community structure of the network in advance. Therefore, we developed a framework, which is generic enough and applicable to all the networks where the ground truth is not necessarily known. In essence, we neither included NMI as the objective function nor took its help in progressing from one generation to another generation. This is a radical and well thought-out departure from the state-of-the-art making our approach in real-life situations.

- We formulated community detection problem as a multi -objective optimization problem with three objectives.

- We proposed two variants: (i) NSGA-III-KRM, we considered Kernel k means, Ratio cut and Modularity as the three objectives, (ii) NSGA-III-CCM, we considered Community fitness, Community score and Modularity as objective functions. We also conducted experiments on two variants of MOEA/D [14] (using the penalty-based boundary intersection method) i.e. MOEA/D-KRM and MOEA/D-CCM with the same parameter combinations and with 20 neighbors.

- We used locus-based representation of community structure to represent a solution. In this, an array of size equal to number of vertices present in the network is used to represent a community structure. It is noteworthy that a single solution can be represented in its various permutations. However, technically all of them are one and the same. Hence, we customized NSGA-III to solve this problem by adding a filter, which checks for the presence of duplicate (permutation) solutions in a generated population at the end of each iteration and if present, they are replaced by a randomly generated solution.

## 5. Basic Deftnitions

### 5.1. *Community Definition*

Community in a network can be described as a subset of nodes that are thickly connected with one another and loosely connected with the remaining nodes present in that network. Intra-links of a given community are represented as the set of edges present inside the community, whereas, interlinks of a given community c are represented by the set of edges connecting the vertices of community c to the vertices not present in community c.

## 5.2. Multi-objective Optimization Problem

Multi-objective optimization problems optimize two or more objective functions simultaneously. Let us consider a problem where we need to maximize *nob* number of objective functions simultaneously as follows:

$$(\max f_1(x)), \max(f_2(x)), \ldots \max(f_{nob}(x))$$

where $x = (x_1, x_2, \ldots x_{noi})$ is the input vector or solutions and $f_1(x), f_2(x), \ldots f_n(x)$ are the objective functions that need to be optimized and *noi* is the dimension of the solution vector. We say that a solution x dominates another solution y, if all the objective functions values with the solution x are better or equal to the respective values of the objective functions with the solution y and at least one objective function value with x is strictly better than the respective objective function value with $x_j$ as input [15]. Else, we say that the solution $x_i$ does not dominate solution $x_j$. We call a solution set S non-dominated if any pair of the solutions present in that set S does not dominate each other.

More than one solution often exists for these types of problems. If we were given a set S with all possible solutions, then the subset of the solution set S i.e. $T_1$ is called Pareto-set with respect to solution set S if it contains all the solutions which do not dominate each other and dominate the rest of the solutions $S - T_1$. Similarly, second Pareto front $T_2$ is the set of solutions, which is subset to set $S-T_1$ which contains all the solutions which do not dominate each other and dominates the rest of the solutions $S-T_1-T_2$. Similarly, third Pareto front, fourth Pareto front etc. are defined.

## 6. Proposed Methodology

### 6.1. Problem Formulation

**First variant:** *Kernel k means, Ratio cut and Modularity* as the objective functions.

$Min\ f_1(x) = Kernel\ K\ Means$

$Min\ f_2(x) = Ratio\ cut$

$Max\ f_3(x) = Modularity$

*Subject to* $x \in X$,

Here vector *x* is a community structure of a network encoded using locus-based representation explained in the next subsection D and X is the set of all possible community structures in a network.

**Second variant**: *Community fitness, Community score and Modularity* as the objective functions.

$Max\ f_1(x) = Community\ Fitnes$

$Max\ f_2(x) = Community\ Score$

$$Max\ f_3(x) = Modularity$$

Subject to $x \in X$,

Here vector *x* is a community structure of a network encoded using locus-based representation explained in the next subsection 6.3 and X is the set of all possible community structures in a network.

### 6.2. Objective functions considered and justification

*Kernel k means* (*KKM*) [16] is used to find dense communities in a network. KKM is computed as follows:

$$KKM = 2(n-m) - \sum_{i=1}^{m} \frac{L(V_i, V_i)}{|V_i|}$$

where n is the number of vertices in a network, m is the number of communities in a network, $|V_i|$ is the number of vertices in community i, $L(V_i, V_i) = \sum_{i,j \in v_i} A_{ij}$ where A is the adjacency matrix of the network. *KKM* should be minimized in order to get structures having denser communities.

*Ratio cut* (*RC*) [17] is used to find the clusters in a network such that each cluster present in it is sparsely connected to the remaining other clusters. The formula for computing the *Ratio cut* is as follows:

$$RC = \sum_{i=1}^{m} \frac{L(V_i, \overline{V_i})}{|V_i|}$$

Where m is the number of communities in a network, $L(V_i, \overline{V_i}) = \sum_{i \in V_i, j \in \overline{V_j}} A_{ij}$ where A is adjacency matrix of the network. Here $\overline{V_i}$ is the set of vertices in the graph but not present in the set $V_i$. *Ratio cut* needs to be minimized in order to get the community structures with less interlinks.

*Community fitness* (CF) [18] is another measure used to find dense communities in a network. When its reaches its highest value, the number of external links is minimized. The formula for computing the CF is as follows:

$$CF = \sum p(s) = \sum_{i=1}^{k} p(s_i)$$

$$where\ p(s) = \sum_{i \in S} \frac{K_i^{in}(s)}{[K_i^{in}(s) + K_i^{out}(s)]^\alpha}$$

where s is the community in a network, $K_i^{in}(s)$ and $K_i^{out}(s)$ are the internal and external degrees of nodes present in the community s, and $\alpha$ is the positive real valued parameter controlling the community size. We considered $\alpha$ value as 1. The higher the value of the parameter, the smaller is the size of the communities found.

*Community score* (*CS*) [5] measures the quality of the division in communities of a network. The higher the *CS*, the denser the clusters obtained. The formula for computing the *CS* is as follows:

$$CS(s) = \sum_{i=1}^{k} score(s_i)$$

$$score(s) = M(s) * V_s$$

$$M(s) = \frac{\sum_{i \in s}(\mu_i)^r}{|s|}$$

$$V_s = \sum_{i,j \in S} A_{ij}$$

$$\mu_i = \frac{1}{|s|} K_i^{in}(s)$$

Where, $\mu_i$ denotes the fraction of edges connecting node $i$ to the other nodes in s, $|s|$ denotes the cardinality of s, S is the set of communities, the exponent r increases the weight of nodes having few connection inside community s. we considered r value as 1 while conducting experiments, score of a community s i.e. $score(s)$ is the product of power mean of s of order r i.e. $M(s)$, and $V_s$, is the volume of the community s, A is the adjacency matrix of the network.

*Modularity* [19] is defined as the fraction of the edges that fall within the given groups minus the expected fraction if the edges were distributed at random. The *Modularity* is computed as follows:

$$Modularity = Q = \sum_{s=1}^{k} \left[ \frac{l_s}{m} - \left( \frac{d_s}{2m} \right)^2 \right]$$

where $l_s$ is the number of intra-links present in community s, $d_s$ is the sum of degrees of nodes in community s, m is the total number of edges in a network, k is the number of communities found inside a network. The greater the *Modularity* value, the desirable is its community structure.

### 6.3. Representation of Solution

The community detection problem formulated as a multi-objective optimization problem, turns out to be a combinatorial optimization problem. Therefore, we need to suitably represent a community, which becomes a solution in the optimization parlance. Toward this end, we used locus based representation taking cue from [20] and [21]. Here, we consider an n dimensional array to represent a solution, where n is the number of nodes in the network. Each cell index in the array represents a node in the network. A cell with label i which represents node i in the network can have value i itself or the labels of nodes which are connected to the node i with an edge in the network. It is to be noted that a single solution can be represented in its various permutations. However, technically all of them are one and the same.

### 6.4. NSGA-III Algorithm

Non-dominated sorting genetic algorithm III (NSGA-III) [2] is a multi and many-objective optimization algorithm and used to optimize three to 15 objective functions simultaneously. This algorithm yields well-diversified and converged solutions. It uses a reference-based framework in order to select a set of solutions from a substantial number of non-dominated solutions to look for diversity. For more details, the reader is referred to [2].

### 6.5. Customizations performed

In this paper, we performed two customizations on the NSGA-III based approach: (i) As a single solution can be represented in various ways (meaning its permutations), in a population for any iteration, if a solution is repeated more than once, then we replace it with a randomly generated solution, (ii) Another customization is that we excluded a solution in which entire network is considered as a single community.

### 6.6. Evaluation Functions

Normalized Mutual Information (*NMI*) and *Modularity* are widely used to figure out the performance of various evolutionary algorithms invoked to detect clusters in any network. *NMI* [22] is used to measure the likenesses between two cluster structures. *NMI* can help us calculate how close the clusters detected by an algorithm and the ground truth cluster structure are. The maximum and minimum values possible for *NMI* are 0 and 1 respectively. Higher the *NMI* value between two cluster structures, higher is their likeness. If the *NMI* value is 1 then it means that both the cluster structures are one and the same. The formula for computing the *NMI* is as follows:

$$NMI(A, B) = \frac{-2 \sum_{i=1}^{R} \sum_{j=1}^{D} C_{ij} \log \left( C_{ij} N / C_i C_j \right)}{\sum_{i=1}^{R} C_i \log \left( C_i / N \right) + \sum_{j=1}^{D} C_j \log \left( C_j / N \right)}$$

where, $C_{ij}$ is the number of nodes appeared in both clusters *i* and *j* present in cluster structures A and B respectively. $C_i(C_j)$ is the number of the elements in cluster *i* (cluster *j*) present in cluster structure A(B), N is the total number of nodes in the network. R(D) is the number of clusters' present in the cluster structure A(B). To make our framework more generic we have not considered NMI of the network or any other evaluation function which requires the knowledge of ground truth community structure as in most of the real-world networks, the ground truth community structure is unknown.

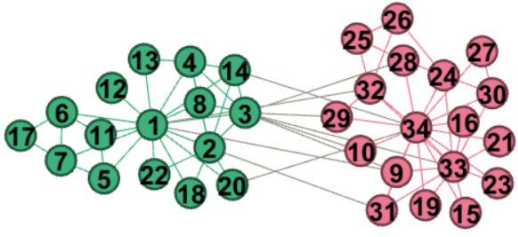

Fig. 1. The D1 network (the ground truth)

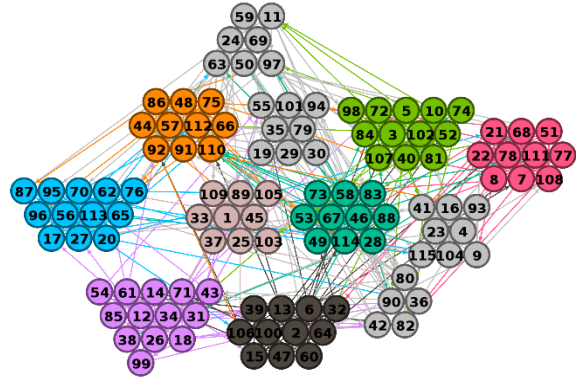

Fig. 3. D3 network (the ground truth).

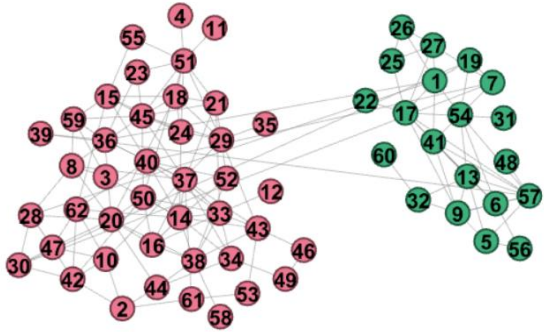

Fig. 2. The D2 network (the ground truth)

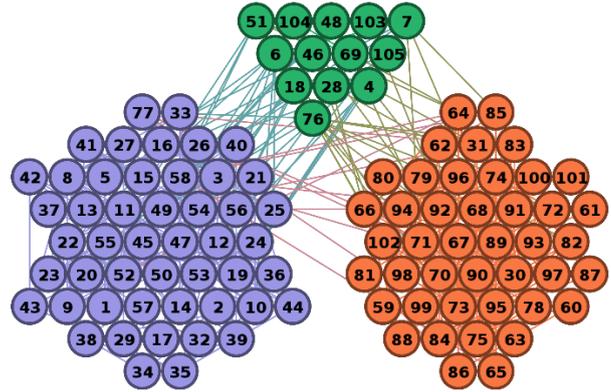

Fig. 4. The D4 network (the ground truth)

## 6.7. *Measures of Convergence and Diversity*

To measure the extent of diversity and the state of convergence of the solutions found by multi and many objective optimization algorithms such as NSGA-III, at the end of a run (in other words, after convergence) two widely used criteria include Inverted Generational Distance (IGD) [2][23] and Hyper volume (HV) [24].

IGD is computed as follows:

$$IGD(A, Z_{eff}) = \frac{1}{|Z_{eff}|} \sum_{i=1}^{|Z_{eff}|} \min_{j=1}^{|A|} d(z_i, a_j)$$

Where, $d(z_i, a_j) = \left\lVert z_i - a_j \right\rVert_2$, A is the set of solutions obtained by the algorithm, $Z_{eff}$ is the set of points present in Pareto optimal surface. $a_j$ is a solution present in set A. $z_i$ is a solution in the Pareto optimal surface which is near to $a_j$.

The IGD measure indicates how close the obtained solutions are to the solutions present in the true Pareto front or Pareto optimal surface. In cases where the true Pareto front is unknown, we run the algorithm by taking large population size and large number of generations. Then, the first Pareto front solutions obtained at the end of the execution are considered as approximation to the Pareto optimal solutions [25]. In our case we considered population size as 500 and number of generations as 500 to approximate Pareto optimal surface.

The Hyper volume [24] of set X is the volume of space formed by non-dominated points present in set X with any reference point. Here the reference point is the "worst possible" point or solution (any point that is dominated by all the points present in solution set X) in the objective space. For a maximization (minimization) problem with positive (negative) valued objectives, we consider origin as the reference point. If a set X has a higher hyper volume than that of a set Y, then we say that X is better than Y.

## 7. Dataset Description

Four benchmark datasets were analyzed in this paper: (i) **Zachary's Karate Club** [26] having 34 nodes and 78 edges with two ground truth communities (Fig. 1) (ii) **Bottlenose Dolphin** [27] with 62 nodes, 159 edges and two ground truth communities (Fig. 2) *(iii) **American College Football** [28] having 115 nodes, 616 edges with twelve ground truth communities (Fig. 3) and finally, (iv)**Books about US Politics** [29] with 105 nodes, 441 edges and three ground truth communities (Fig. 4). Henceforth, we refer the datasets ***Zachary's Karate Club, Bottlenose Dolphin, American College Football** and **Books about US Politics*** to as ***D1, D2, D3*** and ***D4*** respectively for the sake of brevity.

## 8. Experiment Analysis, Results and Discussion

### 8.1. Parameter Setting

We performed sensitivity analysis with the parameter combinations presented in Table III on all datasets using our proposed variants. We conducted 10 runs for each parameter combination. We computed the product of the highest *Modularity* and the highest *NMI* obtained towards the finish of each run and then computed the mean of those products (over 10 runs) for each parameter combination. Any parameter combination producing the highest average product of *NMI* and *Modularity* is considered the best combination. The best parameter combinations obtained for all datasets are presented as follows. It may be mentioned that in problems where the ground truth is unknown, it is impossible to compute *NMI*. Therefore, we recommend decision making based on *Modularity* taking cue from several works in literature.

For the variant NSGA-III-KRM, we varied the population sizes with values 100, 200, 500 and 400; crossover probabilities with values 0.8, 0.85, 0.9 and 0.9 and mutation probabilities with values 1/34, 1/124, 1/230 and 2/105 for the datasets ***D1, D2, D3 and D4*** respectively. For the variant NSGA-III-CCM, we considered the population sizes 20, 200, 450 and 500; crossover probabilities 0.8, 0.85 and 0.9 and mutation probabilities 1/68, 1/62, 1/230 and 2/105 for the datasets ***D1, D2, D3 and D4,*** respectively. The above combinations were obtained by looking for the average highest product of the *NMI* and *Modularity* over 10 runs among all combinations. The parameters of *Community fitness* and *Community score* are kept fixed $\alpha=1$ and $r=1$ respectively.

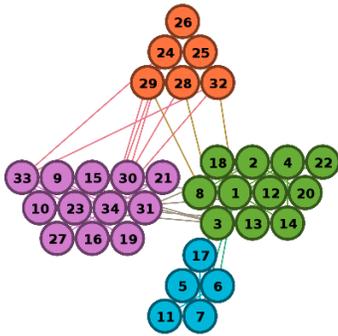

Fig. 5. The obtained clusters of best *Modularity* by NSGA-III-CCM on D1 network.

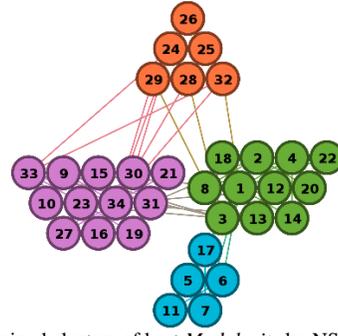

Fig. 7. The obtained clusters of best *Modularity* by NSGA-III -KRM on D1 network.

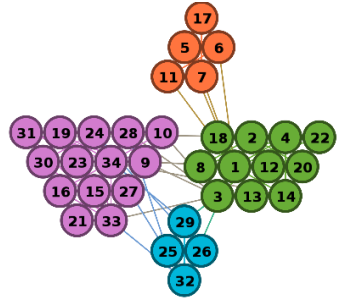

Fig. 6. The obtained clusters of best *NMI* by NSGA-III-CCM on D1 network.

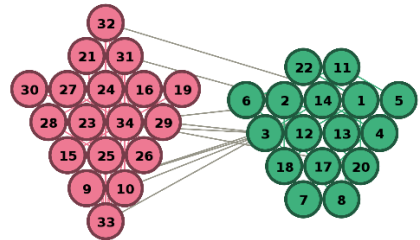

Fig. 8. The obtained clusters of best *NMI* by NSGA-III -KRM on D1 network.

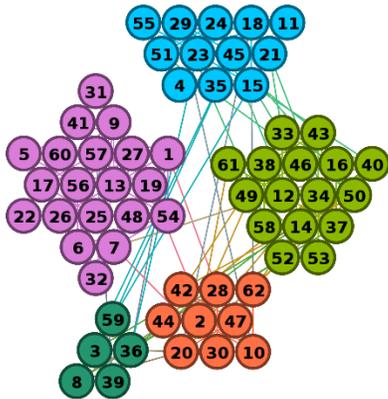

Fig. 9. The obtained communities of best *Modularity* by NSGA-III-CCM on D2 network.

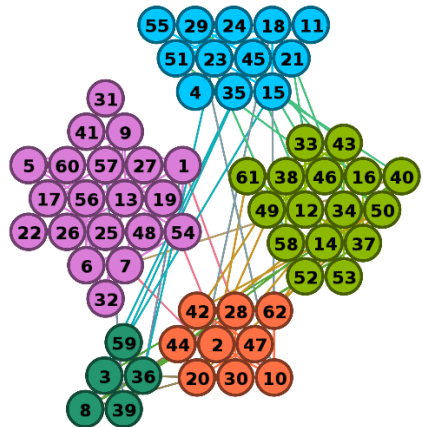

Fig. 11. The obtained communities of best *Modularity* by NSGA-III -KRM on D2 network.

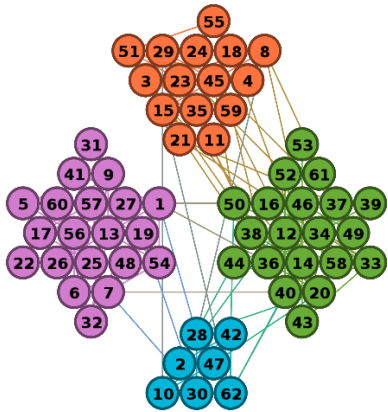

Fig. 10. The obtained communities of best *NMI* by NSGA-III-CCM on D2 network.

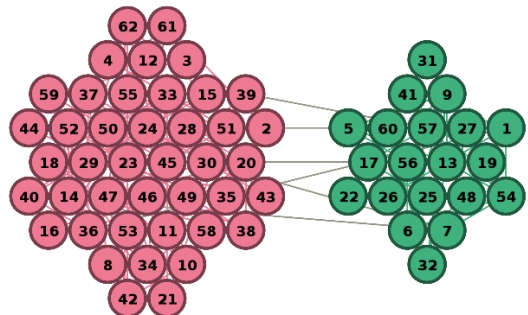

Fig. 12. The obtained communities of best *NMI* by NSGA-III - KRM on D2 network.

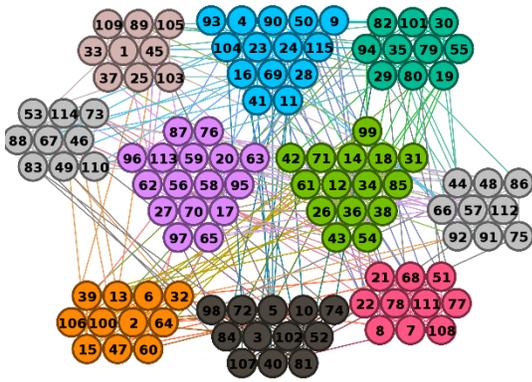

Fig. 13. The obtained community structure by NSGA-III-CCM with highest *Modularity* on D3 network.

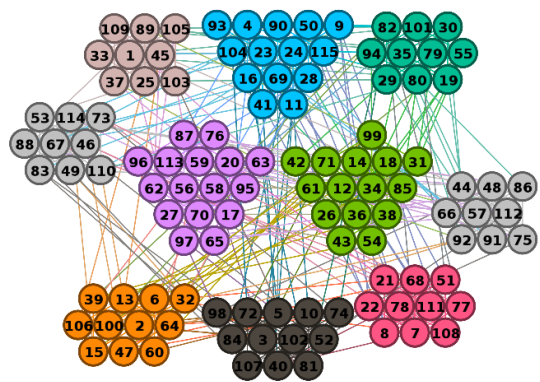

Fig. 15. The obtained community structure by NSGA-III - KRM with highest *Modularity* on D3 network.

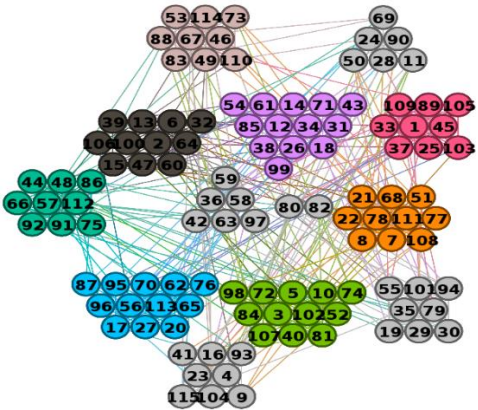

Fig. 14. The obtained community structure by NSGA-III-CCM with highest *NMI* on D3 network.

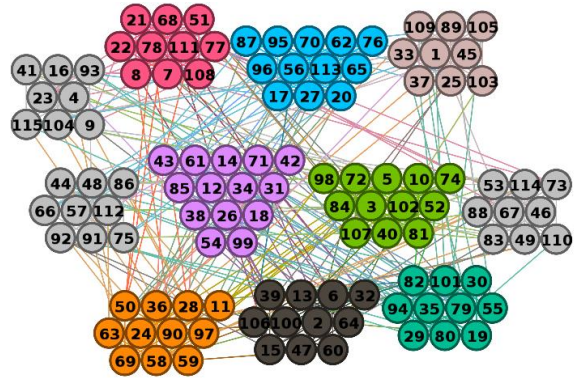

Fig. 16. The obtained community structure of NSGA-III-KRM with highest *NMI* on D3 network.

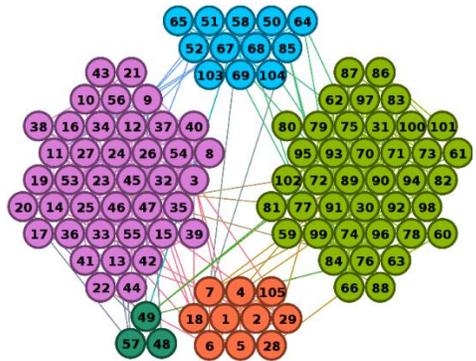

Fig. 17. The obtained community structure by NSGA-III - CCM with highest *Modularity* on D4 network

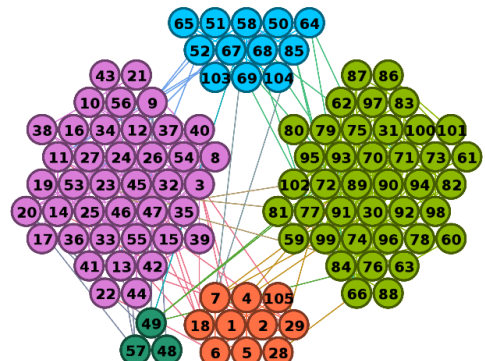

Fig. 19. The obtained community structure by NSGA-III -KRM with highest *Modularity* on D4 network

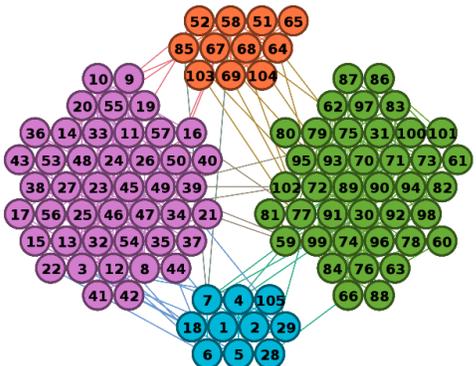

Fig. 18. The obtained community structure by NSGA-III – CCM with highest *NMI* on D4 network

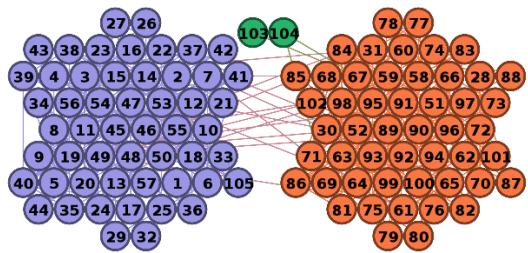

Fig. 20. The obtained community structure by NSGA-III-KRM with highest *NMI* on D4 network

TABLE 1
MAXIMUM AND AVERAGE *MODULARITY* VALUES (QMAX AND QAVG) FOR THE PROPOSED METHODOLOGY FOR 10 RUNS

|    | Index | FN | BGLL | MIGA | Meme-net | GA-net | MOGA-net | MOD-PSO | QIEA-net | iQIEA-net | NSGA-III-CCM | NSGA-III-KRM | MOEA/D–KRM | MOEA/D-CCM |
|----|-------|------|------|------|------|------|------|------|------|------|------|------|------|------|
| D1 | Qmax | 0.3807 | 0.4188 | 0.4188 | 0.402 | 0.4059 | 0.4198 | 0.4198 | 0.4198 | 0.4198 | 0.4198 | 0.4198 | 0.4198 | 0.4198 |
|    | Qavg | 0.3807 | 0.4188 | 0.395 | 0.3855 | 0.4059 | 0.4198 | 0.4182 | 0.4198 | 0.4198 | 0.4198 | 0.4198 | 0.4185 | 0.4167 |
| D2 | Qmax | 0.4897 | 0.5118 | 0.521 | 0.5155 | 0.5014 | 0.5258 | 0.5265 | 0.5213 | 0.5213 | 0.5277 | 0.5285 | 0.521 | 0.5041 |
|    | Qavg | 0.4897 | 0.5118 | 0.4631 | 0.4832 | 0.4948 | 0.5225 | 0.525 | 0.5199 | 0.5211 | 0.5267 | 0.528 | 0.5075 | 0.4873 |
| D3 | Qmax | 0.5497 | 0.6046 | 0.5911 | 0.5888 | 0.594 | 0.528 | 0.6046 | 0.5824 | 0.5988 | 0.6046 | 0.6046 | 0.6046 | 0.601 |
|    | Qavg | 0.5497 | 0.6046 | 0.548 | 0.5432 | 0.5833 | 0.5177 | 0.6015 | 0.5567 | 0.5812 | 0.6038 | 0.6043 | 0.6009 | 0.5971 |
| D4 | Qmax | 0.502 | 0.4986 | 0.4988 | 0.4833 | 0.5033 | 0.4993 | 0.5264 | 0.5214 | 0.5269 | 0.527 | 0.5272 | 0.527 | 0.5112 |
|    | Qavg | 0.502 | 0.4986 | 0.483 | 0.4478 | 0.4997 | 0.4618 | 0.5263 | 0.5209 | 0.5266 | 0.5261 | 0.5257 | 0.525 | 0.4956 |

## 8.2. Results and Discussion

This experiment was conducted on a standalone computer having Intel Xeon(R) CPU E5-2640 v4 2.4 GHz, with 8 cores and 32 GB RAM in Ubuntu 16.04 operating system. For visualizing the optimal communities, we employed Circle Pack layout plugin in Gephi tool (https://gephi.org/). The codes for NSGA-III and MOEA/D are adapted from the website https://github.com/msu-coinlab/pymoo and extended.

TABLE 2
MAXMMUM AND AVERAGE NMI VALUES OVER 10 RUNS

|   |   | Index | D1 | D2 | D3 | D4 |
|---|---|---|---|---|---|---|
| NSGAIII-KRM | | NMI max | 1 | 1 | 0.9341 | 0.7256 |
| | | NMI avg | 1 | 0.9846 | 0.9245 | 0.6017 |
| NSGAIII-CCM | | NMI max | 0.7071 | 0.6455 | 0.9314 | 0.5901 |
| | | NMI avg | 0.6912 | 0.6191 | 0.9291 | 0.5533 |
| MOEA/D-KRM | | NMI max | 1 | 1 | 0.9361 | 0.6114 |
| | | NMI avg | 0.8535 | 0.8891 | 0.9043 | 0.5948 |
| MOEA/D-CCM | | NMI max | 0.7071 | 0.4882 | 0.9363 | 0.5249 |
| | | NMI avg | 0.6697 | 0.4608 | 0.9228 | 0.4701 |

TABLE 3
PARAMETER COMBINATION CONSIDERED FOR DATASETS WHEN DOING SENSITIVITY ANALYSIS

| DATASET | POPULATION SIZE | #GENERATIONS | CROSSOVER PROBABILITIES | MUTATION PROBABILITIES |
|---|---|---|---|---|
| D1 | 100, 150, 200 | 100 | 0.8, 0.85, 0.9 | 1/34, 2/34, 1/(2*34) |
| D2 | 200, 250, 300 | 100 | 0.8, 0.85, 0.9 | 1/62, 2/62, 1/(2*62) |
| D3 | 400, 450, 500 | 100 | 0.8, 0.85, 0.9 | 1/115, 2/115, 1/(2*115) |
| D4 | 400, 450, 500 | 100 | 0.8, 0.85, 0.9 | 1/105, 2/105, 1/(2*105) |

The ground truth communities of ***D1, D2, D3 and D4*** networks are depicted in Figs. 3, 4, 5 and 6, respectively.

Figs. 5 and 6 respectively depict the structures of ***D1*** obtained by the variant NSGA-III-CCM with the highest *Modularity* and the highest *NMI* obtained for the best parameter combination (mentioned in the subsection 8.1). Similarly, Fig. 7 and Fig. 8 respectively depict the structures of ***D1*** obtained by NSGA-III-KRM with the highest *Modularity* and the highest *NMI* for the best parameter combination (mentioned in the subsection 8.1). The community structure with the highest *Modularity* obtained using NSGA-III-CCM and the community structure with the highest Modularity obtained by using NSGA-III-KRM are one and the same. Further, these structures have 4 communities in each of the. Out of these four, two are sub communities of the community present in the ground truth community and other two are sub communities of another community present in the ground truth community structure. Furthermore, The community structure with the highest *NMI* obtained using NSGA-III-KRM turned out to be identical to the ground truth community structure.

The optimal community structure of **D2 network** depicted in Fig. 9 and Fig. 11, with the highest modularity values obtained for the best parameter combination (mentioned in the subsection 8.1) respectively for the two variants turned out to be one and the same. This community structure has five communities. Out of these five, one turned to be the same present in the ground truth and other four are the sub communities of another community present in the ground truth community structure.

The optimal community structure of **D2 network** is depicted in Fig. 10 with the highest *NMI* is obtained for the best parameter combination (mentioned in the subsection 8.1) by using NSGA-III-CCM. Here, one community turned out to be the same one present in the ground truth and other three communities are the sub communities of another community present in the ground truth.

The optimal community structure of **D2 network** is depicted in Fig. 12 with the highest *NMI* is obtained for the best parameter combination (mentioned in the subsection 8.1) by using NSGA-III-KRM. It yielded the same structure as the ground truth community structure.

The optimal community structure of **D3** network depicted in Fig. 13 and Fig. 15 with the highest *Modularity* obtained for the best parameter combination (mentioned in the subsection 8.1) respectively for the two variants turned out to be the same. It has 10 communities. Out of these, 4 turned out to be identical to that in the ground truth, 3 are similar to those in the ground truth but with two or three extra nodes, while the remaining 3 are similar to those in the ground truth with two or three less nodes.

The optimal community structure of **D3** network with the highest *NMI* obtained for the best parameter combination (mentioned in the subsection 8.1) by using NSGA-III-CCM is depicted in Fig. 14. It has 13 communities in it. Out of these, 9 turned out to be identical to the ground truth, 2 are similar as in the ground truth but with one or two less nodes, while the remaining 3 contains nodes of two small communities present in the ground truth.

The optimal community structure of **D3** network with the highest *NMI* obtained for the best parameter combination (mentioned in the subsection 8.1) by using NSGA-III-KRM is depicted in Fig. 16. It contains 11 communities in it. Out of these 11, 6 turned out to be identical to the ones in the ground truth, 3 are similar as in the ground truth but with two or three extra nodes, while the remaining 2 are similar to those in the ground truth but with 1 or 2 less nodes.

The optimal community structure of **D4** network depicted in Fig. 17 and Fig. 19 with the highest *Modularity* obtained for the best parameter combination (mentioned in the subsection

8.1) respectively by using both variants turned out to be identical. It has 5 communities in it. Out of these 5, 2 are sub communities of two communities present in the ground truth having two extra nodes belonging to another communities. Other 3 contains nodes belonging to third community in the ground truth and nodes left out in above two communities.

The optimal community structure of **D4** network with the highest *NMI* obtained for the best parameter combination (mentioned in the subsection 8.1) by using NSGA-III-CCM is depicted in Fig. 18. This community structure has 4 communities in it. Out of these 4, 2 are sub communities of two communities present in the ground truth but with two extra nodes belonging to other communities. Other 3 contain nodes belonging to the third community in the ground truth and nodes left out in above two communities.

The optimal community structure of **D4** network with the highest *NMI* obtained for the best parameter combination (mentioned in the subsection 8.1) by using NSGA-III-KRM is depicted in Fig. 20. This community structure has 3 communities in it. Out of these 3, 2 are the sub communities of two communities present in the ground truth having two extra nodes belonging to another communities. Remaining one contains nodes belonging to the third community in the ground truth and nodes left out in above two communities.

As *Modularity* is widely used for comparison in the literature, we too compared the *Modularity* values yielded by different state-of-the-art approaches in the recently published paper [12] with the optimal Modularity obtained by our methods. This is despite the fact that Modularity as an objective function in both the proposed formulations. This is done for the purpose of comparision only.

Accordingly, in Table I we compared the average *Modularity* and maximum *Modularity* obtained by the proposed variant i.e. NSGA-III-KRM and NSGA-III-CCM with that of 9 state-of-art approaches namely, FN, BGLL, MIGA, MEME-net, MOGA-net, MODPSO, QIEA-net and IQIEA-net and also with MOEA/D variants i.e. MOEA/D-KRM and MOEA/D-CCM.

For the **D1** dataset, our proposed variants of NSGA-III, MOGA-net, QIEA-net and IQIEA-net yielded the same *Modularity* values. For **D2** and **D4 datasets**, our both variants of NSGA-III obtained the highest *Modularity* compared to that of all algorithms. For **D3** dataset, BGLL, our proposed NSGA-III variants and MOEA/D-KRM obtained the highest *Modularity*; the mean *Modularity* values obtained by them are close to each other and higher compared to that of the remaining algorithms. It can be very well seen from the Table I that our proposed NSGA-III variants achieved the best or equal *Modularity* value compared to the remaining approaches. The average *NMI* for all the datasets obtained by both variants using the best parameter combination are presented in the Table II. The communities with the highest *Modularity* obtained by both proposed variants are one and the same, when compared with the ground truth communities. The plots of the sensitivity analysis are depicted in Figs. S. 1 to S. 8 in supplementary material.

We observed from Table I that NSGA-III-KRM outperformed NSGA-III-CCM on two datasets D2 and D3, while producing same result on D1. This is attributed to the more information contained in NSGA-III-KRM vis-a-vis NSGA-III-CCM in that the former obtained communities closer to the ground truth.

However, both variants of NSGA-III outperformed MOEA/D variants i.e. MOEA/D-III-KRM and MOEA/D-CCM on all datasets with respect to average *Modularity*. This is because of the superiority of NSGA-III over MOEA/D in obtaining more diverse and better convergent solutions.

Further, to know the diversity and convergence aspects of the solutions obtained by the proposed methods and to see how close the obtained Pareto front is to the true Pareto front or Pareto optimal surface, we computed the ratio of HV and IGD values of solution set obtained at the end of each run. Then, we computed the average HV/IGD ratios for each parameter combination. The results obtained are presented in the Tables S. I to S. VIII, available in the supplementary material. The ratio HV/IGD is indeed proposed for the first time as a proxy for the empirical attainment function plots used in the bi-objective optimization algorithms because a similar kind of plot is not yet proposed in the literature for multi/many objective optimization algorithms. This is another significant contribution of the study.

## 9. Conclusions

A novel multi-objective community detection framework with two variants i.e. NSGA-III-KRM, NSGA-III-CCM has been proposed in this paper. In the first variant i.e. NSGA-III-KRM, three functions -- *Kernel k means*, *Ratio cut* and *Modularity* – are used as the objective functions. In the second variant, i.e. NSGA-III-CCM, three measures -- *Community fitness*, *Community score* and *Modularity* – are used as the objective functions. A filter has been added in the NSGA-III algorithm which checks for redundant solutions presents in the population at the end of each iteration. The product of *Modularity* and *NMI* is considered to find the best parameter combination. Both proposed variants, NSGA-III-KRM and NSGA-III-CCM, are compared with nine state-of-the-art algorithms and MOEA/D variants (MOEA/D-KRM and MOEA/D-CCM). The results indicate that our proposed variants yielded the best or identical results in terms of *Modularity*. Hence, we conclude that our proposed variants have found community structures in a network with high *Modularity*, indicating that the nodes in the communities are thickly connected with one another and nodes in different communities are well separated, which is a hallmark of this study. We also proposed a new measure, which is an alternative to the empirical attainment function plot available in bi-objective optimization framework.

# I. supplementary material

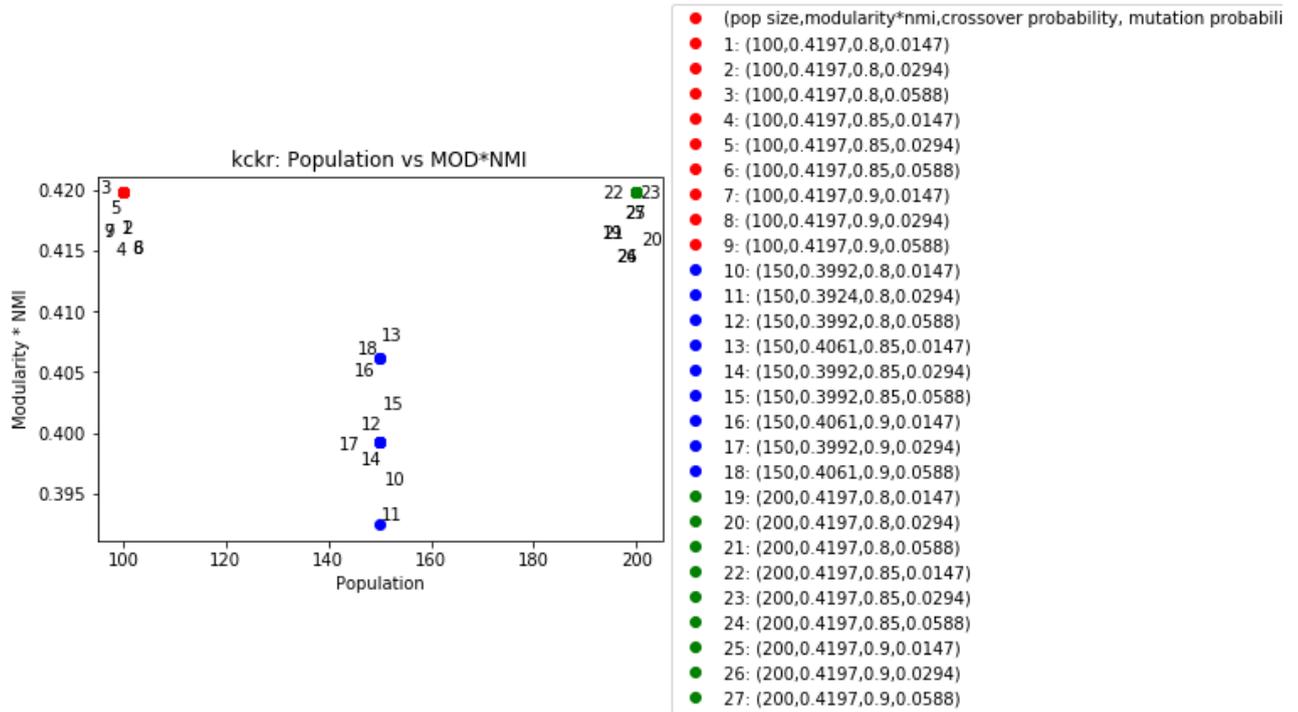

Fig. S. 21. Modularity*NMI vs Population size of karate club dataset using NSGA3-KKM

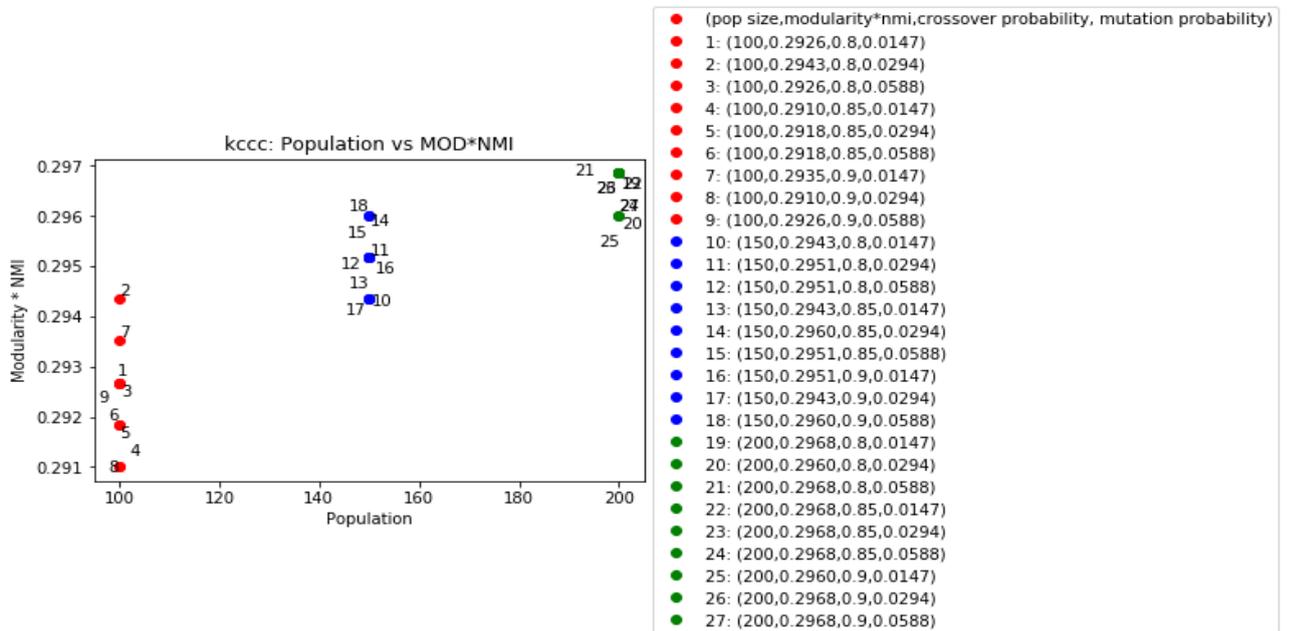

Fig. S. 22. Modularity*NMI vs Population size of Zachary's Karate Club dataset using NSGA3-CCM

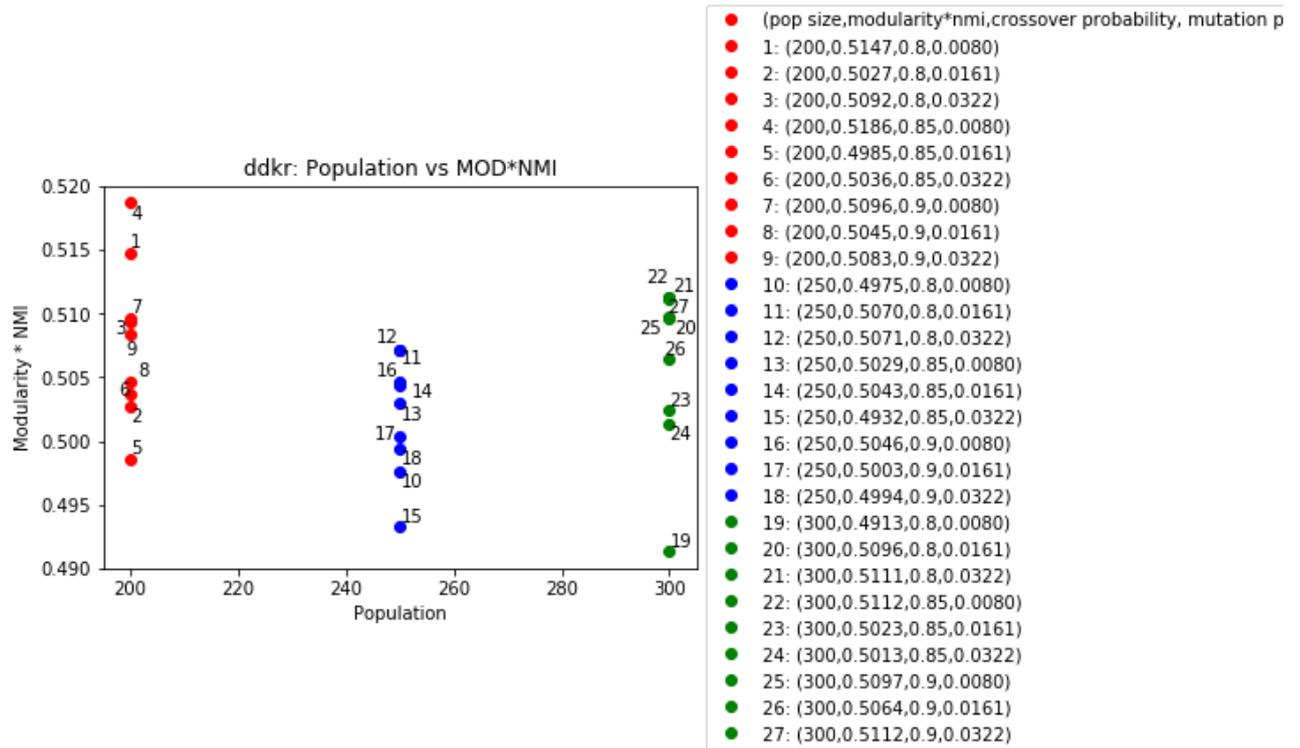

Fig. S. 23. Modularity*NMI vs Population size of Bottlenose Dophin dataset using NSGA3-KKM

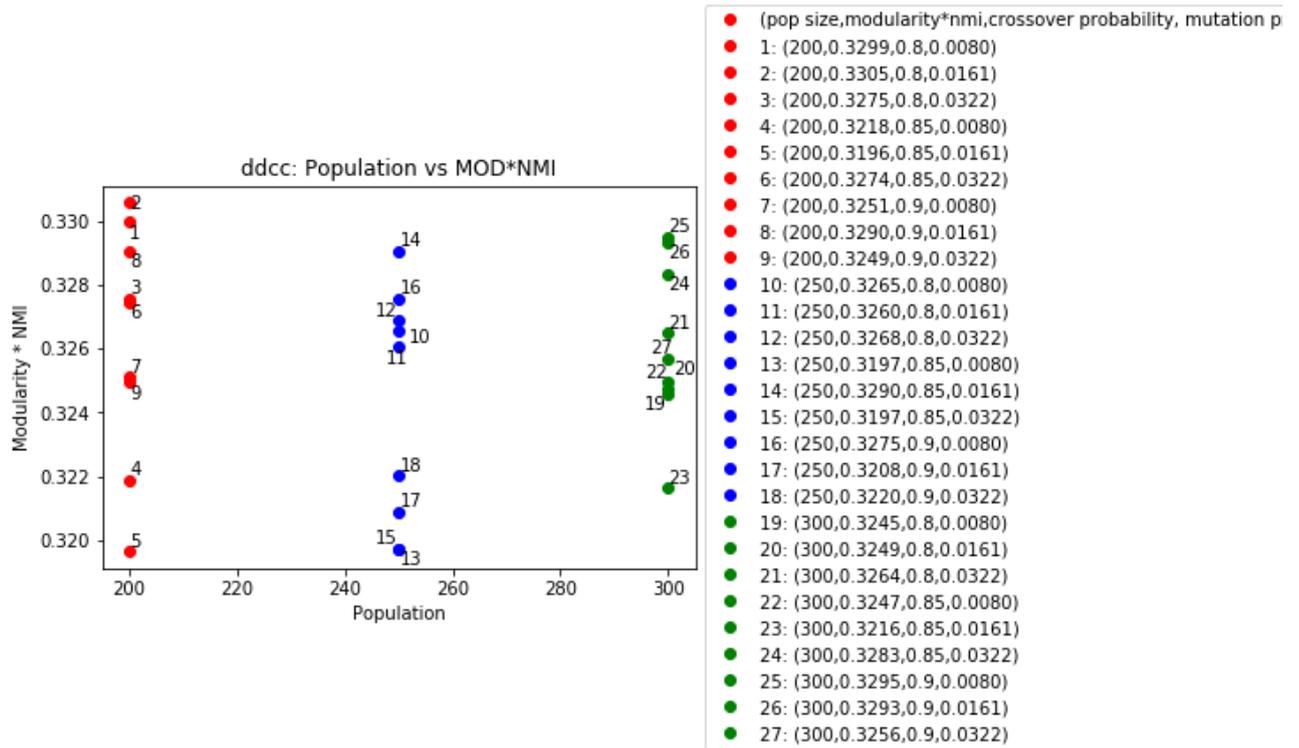

Fig. S. 24. Modularity*NMI vs Population size of Bottlenose Dophin dataset using NSGA3-CCM

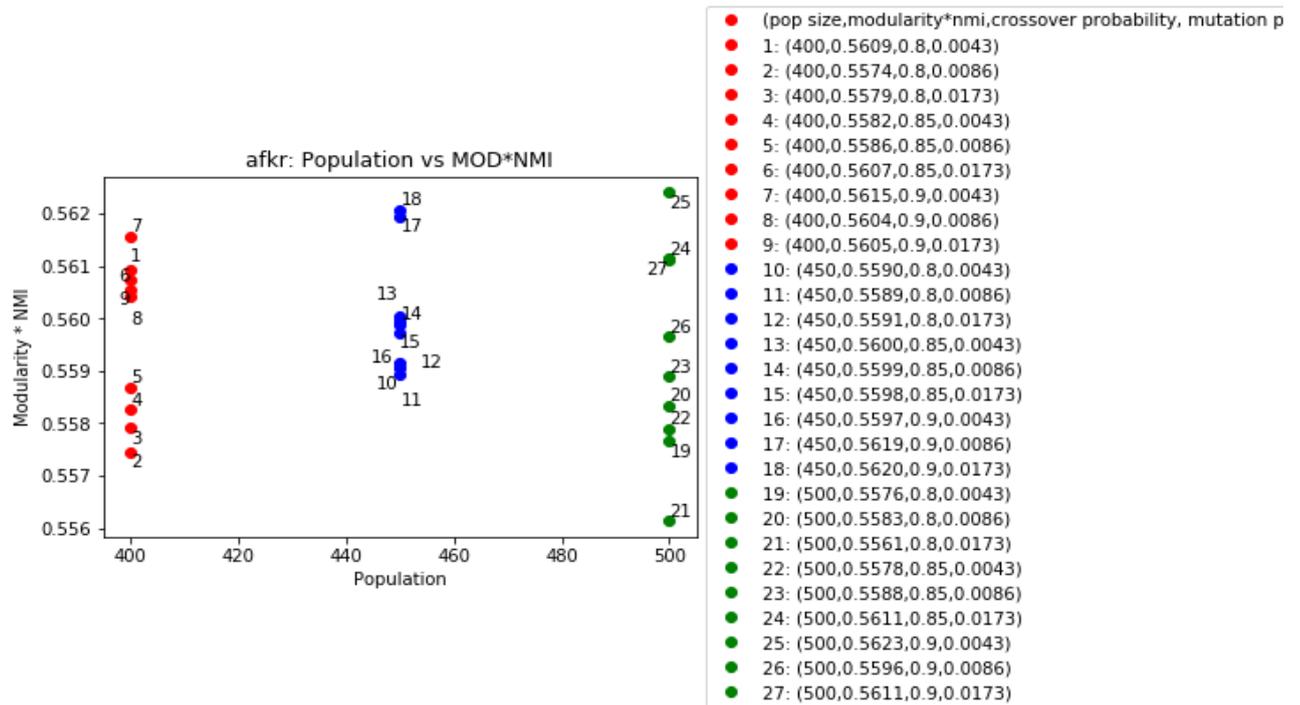

Fig. S. 25. Modularity*NMI vs Population size of American College Football dataset using NSGA3-KKM

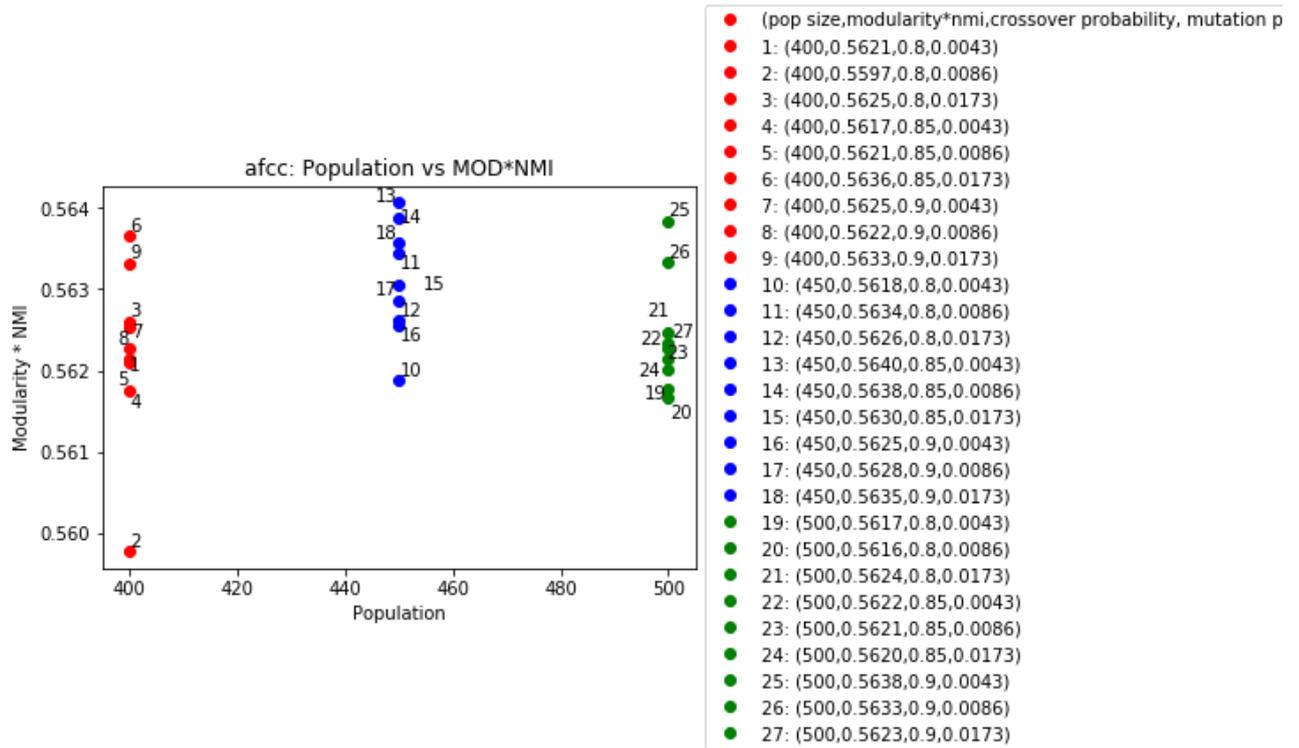

Fig. S. 26. Modularity*NMI vs Population size of American College Football dataset using NSGA3-CCM

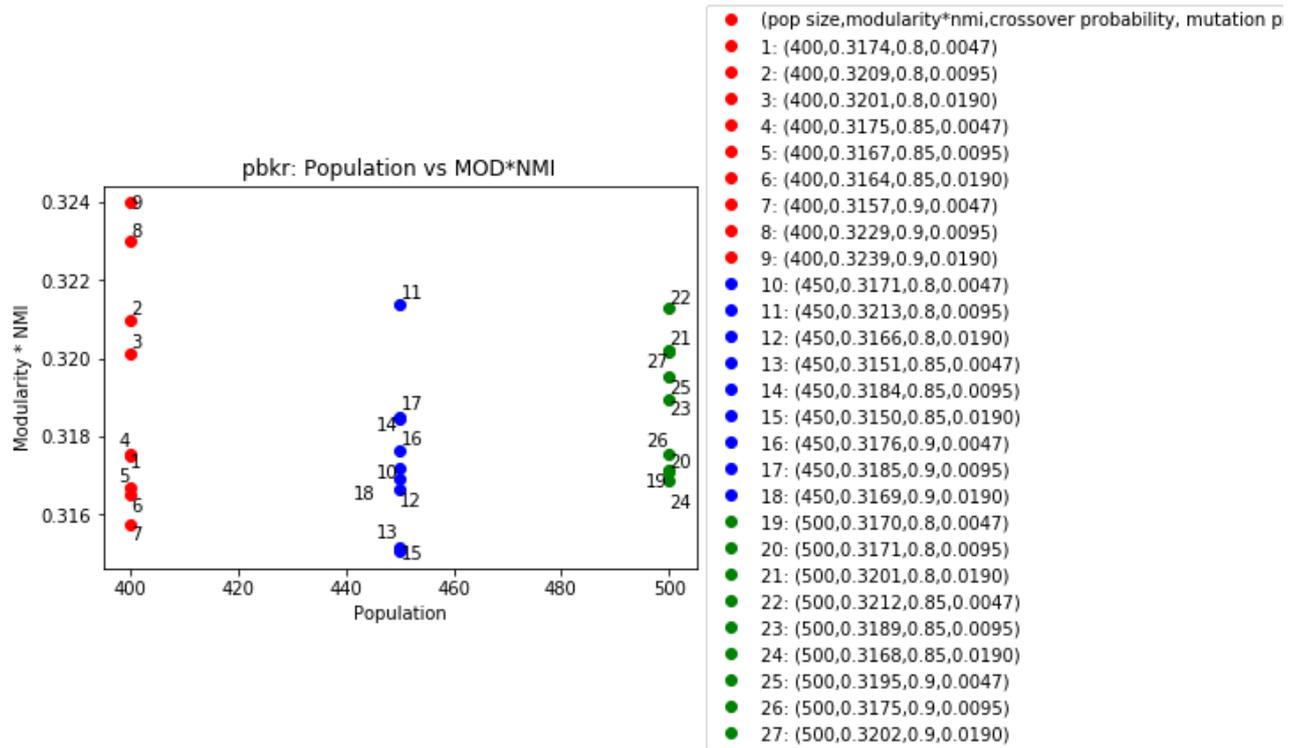

Fig. S. 27. Modularity*NMI vs Population size of Political dataset using NSGA3-KKM

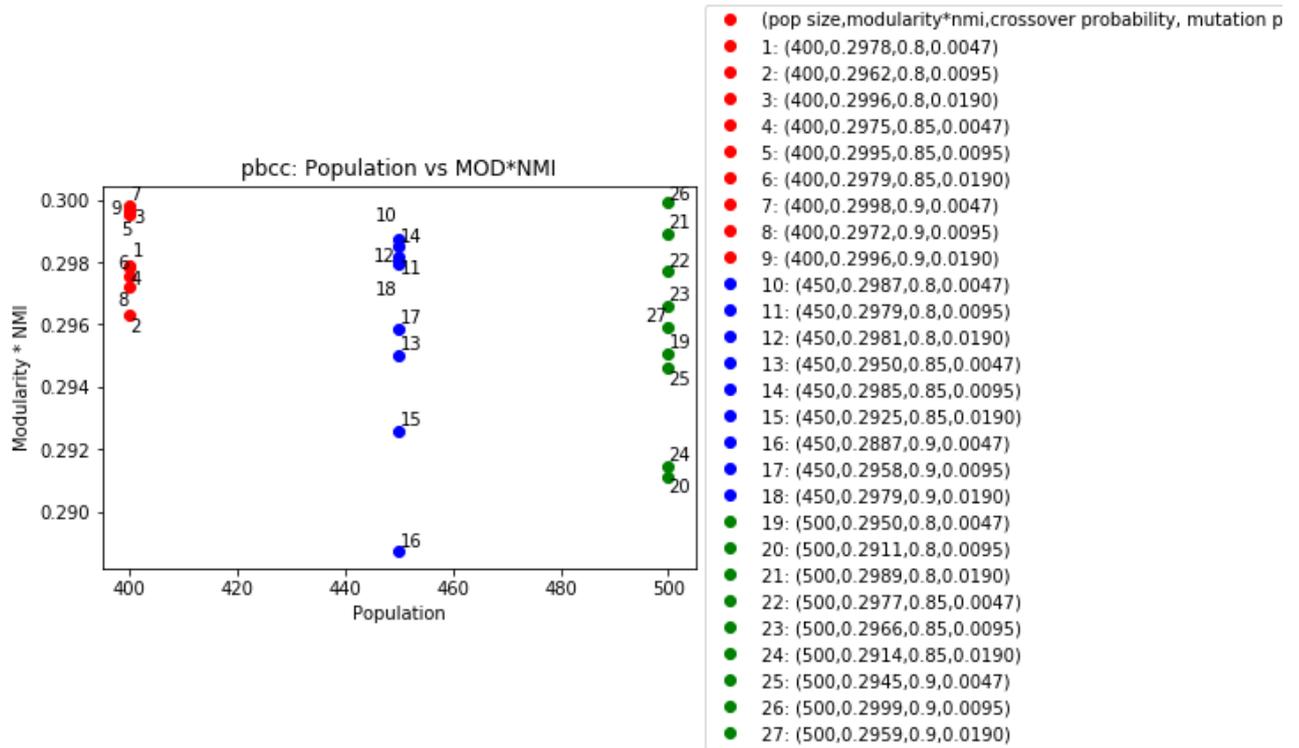

Fig. S. 28. Modularity*NMI vs Population size of Zachary's Karate Club dataset using NSGA3-KKM

TABLE S. I

AVERAGE IGD AND HV VALUES FOR EACH PARAMETER COMBINATION OBTAINED FOR ZACHARY'S KARATE CLUB DATASET USING NSGA III-KRM

| Population size | Generations | Crossover Probability | Mutation Probability | HV/IGD MEAN | HV/IGD MAX |
|---|---|---|---|---|---|
| 100 | 100 | 0.8 | 0.0147 | 130245.6 | 195525.4 |
| 100 | 100 | 0.8 | 0.0294 | 176007 | 530593.9 |
| 100 | 100 | 0.8 | 0.0588 | 119389.5 | 209730 |
| 100 | 100 | 0.85 | 0.0147 | 130076.6 | 199905.9 |
| 100 | 100 | 0.85 | 0.0294 | 152105.7 | 346405.8 |
| 100 | 100 | 0.85 | 0.0588 | 129156.5 | 278482.2 |
| 100 | 100 | 0.9 | 0.0147 | 121333.8 | 281488.7 |
| 100 | 100 | 0.9 | 0.0294 | 125364.7 | 240377.2 |
| 100 | 100 | 0.9 | 0.0588 | 124365.1 | 229433.6 |
| 150 | 100 | 0.8 | 0.0147 | 130245.6 | 195525.4 |
| 150 | 100 | 0.8 | 0.0294 | 176007 | 530593.9 |
| 150 | 100 | 0.8 | 0.0588 | 119389.5 | 209730 |
| 150 | 100 | 0.85 | 0.0147 | 130076.6 | 199905.9 |
| 150 | 100 | 0.85 | 0.0294 | 152105.7 | 346405.8 |
| 150 | 100 | 0.85 | 0.0588 | 129156.5 | 278482.2 |
| 150 | 100 | 0.9 | 0.0147 | 121333.8 | 281488.7 |
| 150 | 100 | 0.9 | 0.0294 | 125364.7 | 240377.2 |
| 150 | 100 | 0.9 | 0.0588 | 124365.1 | 229433.6 |
| 200 | 100 | 0.8 | 0.0147 | 130245.6 | 195525.4 |
| 200 | 100 | 0.8 | 0.0294 | 176007 | 530593.9 |
| 200 | 100 | 0.8 | 0.0588 | 119389.5 | 209730 |
| 200 | 100 | 0.85 | 0.0147 | 130076.6 | 199905.9 |
| 200 | 100 | 0.85 | 0.0294 | 152105.7 | 346405.8 |
| 200 | 100 | 0.85 | 0.0588 | 129156.5 | 278482.2 |
| 200 | 100 | 0.9 | 0.0147 | 121333.8 | 281488.7 |
| 200 | 100 | 0.9 | 0.0294 | 125364.7 | 240377.2 |
| 200 | 100 | 0.9 | 0.0588 | 124365.1 | 229433.6 |

TABLE S. II
AVERAGE IGD AND HV VALUES FOR EACH PARAMETER COMBINATION OBTAINED FOR ZACHARY'S KARATE CLUB DATASET USING NSGA III-CCM

| Population size | Generations | Crossover Probability | Mutation Probability | HV/IGD MEAN | HV/IGD MAX |
|---|---|---|---|---|---|
| 100 | 100 | 0.8 | 0.0147 | 7135.85 | 18741.74 |
| 100 | 100 | 0.8 | 0.0294 | 8183 | 11461.89 |
| 100 | 100 | 0.8 | 0.0588 | 10041.27 | 38391.9 |
| 100 | 100 | 0.85 | 0.0147 | 10191.73 | 21908.42 |
| 100 | 100 | 0.85 | 0.0294 | 7795.31 | 12362.8 |
| 100 | 100 | 0.85 | 0.0588 | 8205.73 | 11545.57 |
| 100 | 100 | 0.9 | 0.0147 | 7741.37 | 15742.72 |
| 100 | 100 | 0.9 | 0.0294 | 11628.64 | 31756.23 |
| 100 | 100 | 0.9 | 0.0588 | 7901.19 | 15167.73 |
| 150 | 100 | 0.8 | 0.0147 | 7135.85 | 18741.74 |
| 150 | 100 | 0.8 | 0.0294 | 8183 | 11461.89 |
| 150 | 100 | 0.8 | 0.0588 | 10041.27 | 38391.9 |
| 150 | 100 | 0.85 | 0.0147 | 10191.73 | 21908.42 |
| 150 | 100 | 0.85 | 0.0294 | 7795.31 | 12362.8 |
| 150 | 100 | 0.85 | 0.0588 | 8205.73 | 11545.57 |
| 150 | 100 | 0.9 | 0.0147 | 7741.37 | 15742.72 |
| 150 | 100 | 0.9 | 0.0294 | 11628.64 | 31756.23 |
| 150 | 100 | 0.9 | 0.0588 | 7901.19 | 15167.73 |
| 200 | 100 | 0.8 | 0.0147 | 7135.85 | 18741.74 |
| 200 | 100 | 0.8 | 0.0294 | 8183 | 11461.89 |
| 200 | 100 | 0.8 | 0.0588 | 10041.27 | 38391.9 |
| 200 | 100 | 0.85 | 0.0147 | 10191.73 | 21908.42 |
| 200 | 100 | 0.85 | 0.0294 | 7795.31 | 12362.8 |
| 200 | 100 | 0.85 | 0.0588 | 8205.73 | 11545.57 |
| 200 | 100 | 0.9 | 0.0147 | 7741.37 | 15742.72 |
| 200 | 100 | 0.9 | 0.0294 | 11628.64 | 31756.23 |
| 200 | 100 | 0.9 | 0.0588 | 7901.19 | 15167.73 |

TABLE S. III

AVERAGE IGD AND HV VALUES FOR EACH PARAMETER COMBINATION OBTAINED FOR BOTTLENOSE DOPHIN CLUB DATASET USING NSGA III-KRM

| Population size | Generations | Crossover Probability | Mutation Probability | HV/IGD MEAN | HV/IGD MAX |
|---|---|---|---|---|---|
| 200 | 100 | 0.8 | 0.0081 | 199009.3 | 316596.1 |
| 200 | 100 | 0.8 | 0.0161 | 229998.6 | 375624.5 |
| 200 | 100 | 0.8 | 0.0322 | 199603 | 302670.6 |
| 200 | 100 | 0.85 | 0.0081 | 192377.3 | 339173.8 |
| 200 | 100 | 0.85 | 0.0161 | 212951 | 348086.3 |
| 200 | 100 | 0.85 | 0.0322 | 188602.5 | 255757.9 |
| 200 | 100 | 0.9 | 0.0081 | 195061.4 | 324119.9 |
| 200 | 100 | 0.9 | 0.0161 | 164325.8 | 248247.1 |
| 200 | 100 | 0.9 | 0.0322 | 186296 | 230404 |
| 250 | 100 | 0.8 | 0.0081 | 199009.3 | 316596.1 |
| 250 | 100 | 0.8 | 0.0161 | 229998.6 | 375624.5 |
| 250 | 100 | 0.8 | 0.0322 | 199603 | 302670.6 |
| 250 | 100 | 0.85 | 0.0081 | 192377.3 | 339173.8 |
| 250 | 100 | 0.85 | 0.0161 | 212951 | 348086.3 |
| 250 | 100 | 0.85 | 0.0322 | 188602.5 | 255757.9 |
| 250 | 100 | 0.9 | 0.0081 | 195061.4 | 324119.9 |
| 250 | 100 | 0.9 | 0.0161 | 164325.8 | 248247.1 |
| 250 | 100 | 0.9 | 0.0322 | 186296 | 230404 |
| 300 | 100 | 0.8 | 0.0081 | 199009.3 | 316596.1 |
| 300 | 100 | 0.8 | 0.0161 | 229998.6 | 375624.5 |
| 300 | 100 | 0.8 | 0.0322 | 199603 | 302670.6 |
| 300 | 100 | 0.85 | 0.0081 | 192377.3 | 339173.8 |
| 300 | 100 | 0.85 | 0.0161 | 212951 | 348086.3 |
| 300 | 100 | 0.85 | 0.0322 | 188602.5 | 255757.9 |
| 300 | 100 | 0.9 | 0.0081 | 195061.4 | 324119.9 |
| 300 | 100 | 0.9 | 0.0161 | 164325.8 | 248247.1 |
| 300 | 100 | 0.9 | 0.0322 | 186296 | 230404 |

TABLE S. IV
AVERAGE IGD AND HV VALUES FOR EACH PARAMETER COMBINATION OBTAINED FOR BOTTLENOSE DOPHIN CLUB DATASET USING NSGA III-CCM

| Population size | Generations | Crossover Probability | Mutation Probability | HV/IGD MEAN | HV/IGD MAX |
|---|---|---|---|---|---|
| 200 | 100 | 0.8 | 0.0081 | 14481.02 | 27483.88 |
| 200 | 100 | 0.8 | 0.0161 | 17151.4 | 24975.29 |
| 200 | 100 | 0.8 | 0.0322 | 11640.36 | 20000.16 |
| 200 | 100 | 0.85 | 0.0081 | 17426.11 | 39490.12 |
| 200 | 100 | 0.85 | 0.0161 | 14118.57 | 25740.14 |
| 200 | 100 | 0.85 | 0.0322 | 17530.78 | 28096.77 |
| 200 | 100 | 0.9 | 0.0081 | 14590.83 | 27520.53 |
| 200 | 100 | 0.9 | 0.0161 | 16103.27 | 28807.43 |
| 200 | 100 | 0.9 | 0.0322 | 18114.06 | 28079.01 |
| 250 | 100 | 0.8 | 0.0081 | 14481.02 | 27483.88 |
| 250 | 100 | 0.8 | 0.0161 | 17151.4 | 24975.29 |
| 250 | 100 | 0.8 | 0.0322 | 11640.36 | 20000.16 |
| 250 | 100 | 0.85 | 0.0081 | 17426.11 | 39490.12 |
| 250 | 100 | 0.85 | 0.0161 | 14118.57 | 25740.14 |
| 250 | 100 | 0.85 | 0.0322 | 17530.78 | 28096.77 |
| 250 | 100 | 0.9 | 0.0081 | 14590.83 | 27520.53 |
| 250 | 100 | 0.9 | 0.0161 | 16103.27 | 28807.43 |
| 250 | 100 | 0.9 | 0.0322 | 18114.06 | 28079.01 |
| 300 | 100 | 0.8 | 0.0081 | 14481.02 | 27483.88 |
| 300 | 100 | 0.8 | 0.0161 | 17151.4 | 24975.29 |
| 300 | 100 | 0.8 | 0.0322 | 11640.36 | 20000.16 |
| 300 | 100 | 0.85 | 0.0081 | 17426.11 | 39490.12 |
| 300 | 100 | 0.85 | 0.0161 | 14118.57 | 25740.14 |
| 300 | 100 | 0.85 | 0.0322 | 17530.78 | 28096.77 |
| 300 | 100 | 0.9 | 0.0081 | 14590.83 | 27520.53 |
| 300 | 100 | 0.9 | 0.0161 | 16103.27 | 28807.43 |
| 300 | 100 | 0.9 | 0.0322 | 18114.06 | 28079.01 |

TABLE S. V
AVERAGE IGD AND HV VALUES FOR EACH PARAMETER COMBINATION OBTAINED FOR AMERICAN COLLEGE FOOTBALL CLUB DATASET USING NSGA III-KRM

| Population size | Generations | Crossover Probability | Mutation Probability | HV/IGD MEAN | HV/IGD MAX |
|---|---|---|---|---|---|
| 400 | 100 | 0.8 | 0.0043 | 439752.6 | 721342 |
| 400 | 100 | 0.8 | 0.0087 | 515214 | 958453.7 |
| 400 | 100 | 0.8 | 0.0174 | 467021 | 633551 |
| 400 | 100 | 0.85 | 0.0043 | 611048.6 | 1389110 |
| 400 | 100 | 0.85 | 0.0087 | 586472.2 | 1130415 |
| 400 | 100 | 0.85 | 0.0174 | 580567.2 | 806541.8 |
| 400 | 100 | 0.9 | 0.0043 | 811402.8 | 1072857 |
| 400 | 100 | 0.9 | 0.0087 | 515076.5 | 1003225 |
| 400 | 100 | 0.9 | 0.0174 | 596012.6 | 1022361 |
| 450 | 100 | 0.8 | 0.0043 | 439752.6 | 721342 |
| 450 | 100 | 0.8 | 0.0087 | 515214 | 958453.7 |
| 450 | 100 | 0.8 | 0.0174 | 467021 | 633551 |
| 450 | 100 | 0.85 | 0.0043 | 611048.6 | 1389110 |
| 450 | 100 | 0.85 | 0.0087 | 586472.2 | 1130415 |
| 450 | 100 | 0.85 | 0.0174 | 580567.2 | 806541.8 |
| 450 | 100 | 0.9 | 0.0043 | 811402.8 | 1072857 |
| 450 | 100 | 0.9 | 0.0087 | 515076.5 | 1003225 |
| 450 | 100 | 0.9 | 0.0174 | 596012.6 | 1022361 |
| 500 | 100 | 0.8 | 0.0043 | 439752.6 | 721342 |
| 500 | 100 | 0.8 | 0.0087 | 515214 | 958453.7 |
| 500 | 100 | 0.8 | 0.0174 | 467021 | 633551 |
| 500 | 100 | 0.85 | 0.0043 | 611048.6 | 1389110 |
| 500 | 100 | 0.85 | 0.0087 | 586472.2 | 1130415 |
| 500 | 100 | 0.85 | 0.0174 | 580567.2 | 806541.8 |
| 500 | 100 | 0.9 | 0.0043 | 811402.8 | 1072857 |
| 500 | 100 | 0.9 | 0.0087 | 515076.5 | 1003225 |
| 500 | 100 | 0.9 | 0.0174 | 596012.6 | 1022361 |

TABLE S. VI

AVERAGE IGD AND HV VALUES FOR EACH PARAMETER COMBINATION OBTAINED FOR AMERICAN COLLEGE FOOTBALL CLUB DATASET USING NSGA III-CCM

| Population size | Generations | Crossover Probability | Mutation Probability | HV/IGD MEAN | HV/IGD MAX |
|---|---|---|---|---|---|
| 400 | 100 | 0.8 | 0.0043 | 60184.29 | 160903.4 |
| 400 | 100 | 0.8 | 0.0087 | 59089.5 | 123946.4 |
| 400 | 100 | 0.8 | 0.0174 | 40253.66 | 85587.95 |
| 400 | 100 | 0.85 | 0.0043 | 46844.43 | 118434.7 |
| 400 | 100 | 0.85 | 0.0087 | 95767.54 | 425939.8 |
| 400 | 100 | 0.85 | 0.0174 | 56949.12 | 133113.8 |
| 400 | 100 | 0.9 | 0.0043 | 44035.57 | 88556.5 |
| 400 | 100 | 0.9 | 0.0087 | 57628.37 | 103291.5 |
| 400 | 100 | 0.9 | 0.0174 | 50456.42 | 132127.6 |
| 450 | 100 | 0.8 | 0.0043 | 60184.29 | 160903.4 |
| 450 | 100 | 0.8 | 0.0087 | 59089.5 | 123946.4 |
| 450 | 100 | 0.8 | 0.0174 | 40253.66 | 85587.95 |
| 450 | 100 | 0.85 | 0.0043 | 46844.43 | 118434.7 |
| 450 | 100 | 0.85 | 0.0087 | 95767.54 | 425939.8 |
| 450 | 100 | 0.85 | 0.0174 | 56949.12 | 133113.8 |
| 450 | 100 | 0.9 | 0.0043 | 44035.57 | 88556.5 |
| 450 | 100 | 0.9 | 0.0087 | 57628.37 | 103291.5 |
| 450 | 100 | 0.9 | 0.0174 | 50456.42 | 132127.6 |
| 500 | 100 | 0.8 | 0.0043 | 60184.29 | 160903.4 |
| 500 | 100 | 0.8 | 0.0087 | 59089.5 | 123946.4 |
| 500 | 100 | 0.8 | 0.0174 | 40253.66 | 85587.95 |
| 500 | 100 | 0.85 | 0.0043 | 46844.43 | 118434.7 |
| 500 | 100 | 0.85 | 0.0087 | 95767.54 | 425939.8 |
| 500 | 100 | 0.85 | 0.0174 | 56949.12 | 133113.8 |
| 500 | 100 | 0.9 | 0.0043 | 44035.57 | 88556.5 |
| 500 | 100 | 0.9 | 0.0087 | 57628.37 | 103291.5 |
| 500 | 100 | 0.9 | 0.0174 | 50456.42 | 132127.6 |

TABLE S. VII
AVERAGE IGD AND HV VALUES FOR EACH PARAMETER COMBINATION OBTAINED FOR BOOKS ABOUT US POLITICS DATASET USING NSGA III-KRM

| Population size | Generations | Crossover Probability | Mutation Probability | HV/IGD MEAN | HV/IGD MAX |
|---|---|---|---|---|---|
| 400 | 100 | 0.8 | 0.0048 | 273578.6 | 519785.3 |
| 400 | 100 | 0.8 | 0.0095 | 336682.7 | 568148.3 |
| 400 | 100 | 0.8 | 0.0191 | 317074 | 606453.4 |
| 400 | 100 | 0.85 | 0.0048 | 315524.4 | 579135.5 |
| 400 | 100 | 0.85 | 0.0095 | 286304.3 | 564227.6 |
| 400 | 100 | 0.85 | 0.0191 | 382656.3 | 529668.8 |
| 400 | 100 | 0.9 | 0.0048 | 321926 | 460325 |
| 400 | 100 | 0.9 | 0.0095 | 288575.4 | 461571.9 |
| 400 | 100 | 0.9 | 0.0191 | 280119.6 | 383115.5 |
| 450 | 100 | 0.8 | 0.0048 | 273578.6 | 519785.3 |
| 450 | 100 | 0.8 | 0.0095 | 336682.7 | 568148.3 |
| 450 | 100 | 0.8 | 0.0191 | 317074 | 606453.4 |
| 450 | 100 | 0.85 | 0.0048 | 315524.4 | 579135.5 |
| 450 | 100 | 0.85 | 0.0095 | 286304.3 | 564227.6 |
| 450 | 100 | 0.85 | 0.0191 | 382656.3 | 529668.8 |
| 450 | 100 | 0.9 | 0.0048 | 321926 | 460325 |
| 450 | 100 | 0.9 | 0.0095 | 288575.4 | 461571.9 |
| 450 | 100 | 0.9 | 0.0191 | 280119.6 | 383115.5 |
| 500 | 100 | 0.8 | 0.0048 | 273578.6 | 519785.3 |
| 500 | 100 | 0.8 | 0.0095 | 336682.7 | 568148.3 |
| 500 | 100 | 0.8 | 0.0191 | 317074 | 606453.4 |
| 500 | 100 | 0.85 | 0.0048 | 315524.4 | 579135.5 |
| 500 | 100 | 0.85 | 0.0095 | 286304.3 | 564227.6 |
| 500 | 100 | 0.85 | 0.0191 | 382656.3 | 529668.8 |
| 500 | 100 | 0.9 | 0.0048 | 321926 | 460325 |
| 500 | 100 | 0.9 | 0.0095 | 288575.4 | 461571.9 |
| 500 | 100 | 0.9 | 0.0191 | 280119.6 | 383115.5 |

TABLE S. VIII
AVERAGE IGD AND HV VALUES FOR EACH PARAMETER COMBINATION OBTAINED FOR BOOKS ABOUT US POLITICS DATASET USING NSGA III-CCM

| Population size | Generations | Crossover Probability | Mutation Probability | HV/IGD MEAN | HV/IGD MAX |
|---|---|---|---|---|---|
| 400 | 100 | 0.8 | 0.0048 | 16709.43 | 30704.67 |
| 400 | 100 | 0.8 | 0.0095 | 17396.99 | 32342.02 |
| 400 | 100 | 0.8 | 0.0191 | 10238.8 | 19612.07 |
| 400 | 100 | 0.85 | 0.0048 | 14768.05 | 34681.02 |
| 400 | 100 | 0.85 | 0.0095 | 16828.09 | 32870.13 |
| 400 | 100 | 0.85 | 0.0191 | 12010.52 | 21758.29 |
| 400 | 100 | 0.9 | 0.0048 | 15860.12 | 33398.29 |
| 400 | 100 | 0.9 | 0.0095 | 12166.83 | 24472.24 |
| 400 | 100 | 0.9 | 0.0191 | 12474.38 | 32580.1 |
| 450 | 100 | 0.8 | 0.0048 | 16709.43 | 30704.67 |
| 450 | 100 | 0.8 | 0.0095 | 17396.99 | 32342.02 |
| 450 | 100 | 0.8 | 0.0191 | 10238.8 | 19612.07 |
| 450 | 100 | 0.85 | 0.0048 | 14768.05 | 34681.02 |
| 450 | 100 | 0.85 | 0.0095 | 16828.09 | 32870.13 |
| 450 | 100 | 0.85 | 0.0191 | 12010.52 | 21758.29 |
| 450 | 100 | 0.9 | 0.0048 | 15860.12 | 33398.29 |
| 450 | 100 | 0.9 | 0.0095 | 12166.83 | 24472.24 |
| 450 | 100 | 0.9 | 0.0191 | 12474.38 | 32580.1 |
| 500 | 100 | 0.8 | 0.0048 | 16709.43 | 30704.67 |
| 500 | 100 | 0.8 | 0.0095 | 17396.99 | 32342.02 |
| 500 | 100 | 0.8 | 0.0191 | 10238.8 | 19612.07 |
| 500 | 100 | 0.85 | 0.0048 | 14768.05 | 34681.02 |
| 500 | 100 | 0.85 | 0.0095 | 16828.09 | 32870.13 |
| 500 | 100 | 0.85 | 0.0191 | 12010.52 | 21758.29 |
| 500 | 100 | 0.9 | 0.0048 | 15860.12 | 33398.29 |
| 500 | 100 | 0.9 | 0.0095 | 12166.83 | 24472.24 |
| 500 | 100 | 0.9 | 0.0191 | 12474.38 | 32580.1 |